\documentclass[letterpaper]{article} 
\usepackage{aaai23}
\usepackage{times}  
\usepackage{helvet}  
\usepackage{courier}  
\usepackage[hyphens]{url}  
\usepackage{graphicx} 
\usepackage{natbib}  
\usepackage{caption} 
\usepackage{algorithm}
\usepackage{algorithmic}
\usepackage{titling}

\nocopyright

\title{Bayesian Inverse Reinforcement Learning for Non-Markovian Rewards}
\date{}
\author{
    Noah Topper\thanks{University of Central Florida},
    Alvaro Velasquez\thanks{University of Colorado Boulder}, George Atia\protect\footnotemark[1]
}

\usepackage{amsmath}
\usepackage{tikz}
\usetikzlibrary{
  patterns,
  calc,
  angles,
  quotes,
  automata,
  positioning,
  decorations.markings,
  arrows,
  intersections,
  shapes,
  arrows,
  backgrounds
}
\newtheorem{definition}{Definition}

\begin{document}
\maketitle
\begin{abstract}
    Inverse reinforcement learning (IRL) is the problem of inferring a reward function from expert behavior. There are several approaches to IRL, but most are designed to learn a Markovian reward. However, a reward function might be non-Markovian, depending on more than just the current state, such as a reward machine (RM). Although there has been recent work on inferring RMs, it assumes access to the reward signal, absent in IRL. We propose a Bayesian IRL (BIRL) framework for inferring RMs directly from expert behavior, requiring significant changes to the standard framework. We define a new reward space, adapt the expert demonstration to include history, show how to compute the reward posterior, and propose a novel modification to simulated annealing to maximize this posterior. We demonstrate that our method performs well when optimizing according to its inferred reward and compares favorably to an existing method that learns exclusively binary non-Markovian rewards.
\end{abstract}

\section{Introduction}
Inverse reinforcement learning (IRL) has become an area of increasing interest in the last decade. Given an expert demonstration, the goal is to infer the reward function that the expert is optimizing. We are typically given a reward-free Markov Decision Process (MDP) and a demonstration in the form of a set of state-action pair trajectories. We then infer a reward function over the MDP.

IRL is fundamentally an ill-posed problem; there are many possible reward functions that yield the same optimal behavior. As such, there are various methods for settling this ambiguity, including margin optimization \cite{Ng00algorithmsfor} \cite{Abbeel04apprenticeshiplearning} \cite{Neu2007}, maximum entropy \cite{ziebart2008maximum} \cite{Wulf2015} \cite{Boularias2012}, and Bayesian \cite{Ramachandran2007} \cite{Lopes09activelearning} \cite{NIPS2011_c51ce410} methods. We focus on the Bayesian approach.

There has also been recent interest in non-Markovian RL using reward machines (RMs) \cite{ToroIcarte2020}. In such settings, the RL agent must optimize for a non-Markovian reward, meaning one that is history-dependent. RMs are a certain finite-state machine representation of such memory-based reward functions.

There has even been recent work in \emph{inferring} these RMs by observing the environment \cite{Icarte2021} \cite{Xu2019} \cite{Xu2020}. However, such methods must be able to observe the reward signal. We propose a method for inferring an RM from expert behavior \emph{only}.

In this paper, we extend the standard Bayesian IRL (BIRL) framework to the non-Markovian setting using RMs and develop a novel simulated annealing solution. We then evaluate our algorithm on a testbed of gridworlds and compare against an exiting algorithm which learns exclusively binary non-Markovian rewards.

\section{Preliminaries}
\label{sec:prelim}
\begin{definition} A \emph{reward-free Markov Decision Process $M$} is a tuple $(S, s_I, A, T)$ with $S$ a finite set of states, $s_I$ the initial state, $A$ a finite set of actions, and $T: S \times A \times S \to [0, 1]$ a probabilistic transition function.
\end{definition}

In standard IRL, we are given MDP $M$ and demonstration $\mathcal{D}$, and must infer a reward function $R: S \to \mathbf{R}$. We assume $\mathcal{D}$ is produced by some expert, and we wish to infer an $R$ that best explains its behavior.

In the BIRL framework, we define a prior $P(R)$ over possible reward functions and $\mathcal{D}$ is given as a set of $N$ state-action pairs, $\mathcal{D} = \{(s_i, a_i) \mid 1 \leq i \leq N\}$. Let $Q(s, a, R)$ denote the optimal Q-value of $(s, a)$ in $M$ given $R$. We then assume the expert behaves according to the Boltzmann rationality model: $P(a \mid s, R) = e^{\alpha Q(s, a, R)}/\Sigma_{a'} e^{\alpha Q(s, a', R)}$.


This is a sort of softmax over action values with rationality factor $\alpha$, where actions with higher long-term reward are exponentially more likely. The higher $\alpha$ is, the closer the expert is to optimal. Assuming $\mathcal{D}$ is a single trajectory, we can compute its probability as:
\begin{equation}
    \begin{split}
        P(\mathcal{D} \mid R) &= \prod_i P(a_i \mid s_i, R) P(s_i \mid s_{i-1}, a_{i-1}) \\
                              &= \prod_i \frac{e^{\alpha Q(s_i, a_i, R)}}{\Sigma_{a} e^{\alpha Q(s_i, a, R)}} \prod_i P(s_i \mid s_{i-1}, a_{i-1}) \\
                              &= \frac{e^{\alpha \Sigma_i Q(s_i, a_i, R)}}{\Pi_i \Sigma_{a} e^{\alpha Q(s_i, a, R)}} \prod_i P(s_i \mid s_{i-1}, a_{i-1}). \\
    \end{split}
\end{equation}

The product term on the right is simply the probability of observing the demonstrated state sequence, given the action sequence. This is determined by the dynamics and does not vary with $R$, so we can treat it as a constant. If $\mathcal{D}$ contains multiple trajectories, only this right term will be affected, and will still be constant with respect to $R$. Thus we can in general say: $P(\mathcal{D} \mid R) \propto e^{\alpha \Sigma_i Q(s_i, a_i, R)}/\Pi_i \Sigma_{a} e^{\alpha Q(s_i, a, R)}$.


We can then compute the reward posterior as $P(R \mid \mathcal{D}) \propto P(\mathcal{D} \mid R)P(R)$, and we can find the likelihood ratio between any two reward functions, which is all we need.

\citet{Ramachandran2007} showed that the mean of the reward posterior is a good estimator, in the sense that optimizing with respect to it will minimize expected policy loss. They use Markov chain Monte Carlo (MCMC) to sample from the posterior and average many such samples.

To estimate the shape of a distribution using MCMC, start at a random point, generate a random neighbor, transition to this neighbor according to some probabilistic rule, and repeat. With a well-chosen rule, we will visit each point proportional to its posterior probability.

\citet{Ramachandran2007} discretize the space of reward vectors and transition from $R_1$ to $R_2$ with probability: $\min \left(1, \frac{P(R_2 \mid \mathcal{D})}{P(R_1 \mid \mathcal{D})}\right) = \min \left(1, \frac{P(\mathcal{D} \mid R_2)P(R_2)}{P(\mathcal{D} \mid R_1)P(R_1)}\right)$. If $R_2$ is more likely than $R_1$, we transition with probability $1$. If not, we still want to accept it sometimes to explore, so we move with probability equal to their likelihood ratio.

\section{Methodology}
\label{sec:method}
\subsection{Non-Markovian BIRL Framework}
In our setting, the reward is a function of histories rather than MDP states. We label our MDP states with relevant observations, and the reward is a function over label sequences.

\begin{definition} A \emph{labeled reward-free Markov Decision Process $M$} is a tuple $(S, s_I, A, T, \Sigma, L)$ with $S$ a finite set of states, $s_I$ the initial state, $A$ a finite set of actions, $T: S \times A \times S \to [0, 1]$ a probabilistic transition function, $\Sigma$ a finite alphabet, and $L: S \to \Sigma$ a state-labeling function.
\end{definition}

\begin{definition}
A \emph{reward machine} $R$ is a tuple $(Y, y_I, \Sigma, \Gamma, \tau, \varrho)$ with $Y$ a finite set of states, $y_I$ the initial state, $\Sigma$ a finite input alphabet, $\Gamma \subset \mathbf{R}$ a finite set of rewards, $\tau: Y \times \Sigma \to Y$ a transition function, and $\varrho: Y \times \Sigma \to \Gamma$ a reward function.
\end{definition}

Note that we are assuming a finite set of possible reward values. We will take this set as known during inference.

\begin{definition}
Given a labeled reward-free MDP $M = (S, s_I, A, T, \Sigma, L)$ and RM $R = (Y, y_I, \Sigma, \Gamma, \tau, \varrho)$, the \emph{product} $M \times R$ is a standard MDP $(S', s_I', A, T', R')$ where:
\begin{itemize}
    \item $S' = S \times Y$
    \item $s_I' = (s_I, y_I)$
    \item $T'((s, y), a, (s', y')) = \\
        \begin{cases}
            T(s, a, s') & \text{if } \tau(y, L(s')) = y', \\
            0 & \text{otherwise}
        \end{cases}$
    \item $R'((s, y), a, (s', y')) = \varrho(y, L(s'))$   
\end{itemize}
\end{definition}

Intuitively, we take $M$ and $R$ together to get a standard MDP. For $M \times R$, we take actions in $M$ and feed the resulting labels into $R$ to generate rewards. Rewards then depend only on the joint state of the system and are thus Markovian. We assume the expert is optimizing such a joint MDP, but the learner only has access to $M$ and $\mathcal{D}$. To estimate $R$, we must depart from the classic BIRL framework in a number of important ways.

Firstly, $\mathcal{D}$ must include more than just state-action pairs. We are inferring a non-Markovian reward function, so knowing which actions the expert took from which MDP states is insufficient; we also need to know the relevant history.

Let $\mathcal{D} = \{(s_i, \lambda_i, a_i) \mid 1 \leq i \leq N\}$, where $\lambda_i \in \Sigma^*$ is the label sequence the expert observed up to the point of taking action $a_i$. Given RM $R$, we can take the product $M \times R$ to get a standard MDP. We then project each $(s_i, \lambda_i, a_i)$ onto the state space of $M \times R$, where $\lambda_i$ induces a specific state in $R$ by stepping through the hypothesis according to $\tau$.

Thus for each hypothesis $R$ we have a standard MDP and a set of state-action pairs $\mathcal{D}$. We can now use BIRL to compute the posterior for $R$. In $M \times R$, let $Q(s, \lambda, a, R)$ denote the optimal Q-value of taking action $a$ from state $s$, having observed trace $\lambda$. The posterior is computed similar to before, with $P(R\mid\mathcal{D}) \propto P(\mathcal{D} \mid R)P(R)$, where now $P(\mathcal{D} \mid R) \propto e^{\alpha \Sigma_i Q(s_i, \lambda_i, a_i, R)}/\Pi_i \Sigma_{a} e^{\alpha Q(s_i, \lambda_i, a, R)}$.


Next, even if we can sample RMs from this posterior, it is not clear what it would mean to average them. Instead, we seek a MAP estimator. Simulated annealing is a method similar to MCMC, but which seeks to maximize some distribution, rather than estimate its whole shape. As such, it naturally fits our setting. Let us summarize the process.

Let $\mathcal{R}_n(\Sigma, \Gamma)$ be the space of RMs with $n$ states over alphabet $\Sigma$ and reward set $\Gamma$. Let $m = |\Sigma|$. An element of $\mathcal{R}_n$ is a tuple $(Y_n, t, r)$ where $Y_n = \{1, ..., n\}$ is the state space (assume $1$ is the initial state), $t \in Y^{n \times m}_n$ is a transition matrix, and $r \in \Gamma^{n \times m}$ is a reward matrix.

Fix $n$ as known. Choose an initial point $R_1$ by setting the matrices uniformly at random. Then iteratively propose a ``neighboring'' RM $R_2$ and transition to it with probability: $\min \left(1, \left(\frac{P(\mathcal{D}\mid R_2)P(R_2)}{P(\mathcal{D}\mid R_1)P(R_1)}\right)^{1 / T_i}\right)$, with ``temperature'' $T_i$.


This is equivalent to simulated annealing with energy equal to the log posterior of the RMs. With the proper temperature schedule and choice of neighbors, we will in principle eventually converge to the peak of the distribution.

Now we consider how to propose neighbors. In early experiments, we tried perturbing a single transition, altering one entry of either the $t$ or $r$ matrix to a new random state or reward. This proved to be too little change. We often need to change multiple factors at once to escape a local optimum.

Instead, fix a probability $p$. Each hypothesis transition (i.e. each entry of $t$ and $r$) is, with independent probability $p$, changed to a new uniformly random value, with a minimum of one change. In fact, we use a probability $p_i$, gradually decaying the value along with the temperature. This allows us to make large changes early on but gradually slow down as we narrow in on a better region of the hypothesis space.

As a technical detail, we restrict attention to RMs in which all states are reachable from the start state. This prevents ``repeat'' hypotheses from showing up at different sizes. We also consider the ``trivial'' RM, one with all zero rewards, to be invalid. If such an RM is sampled, we simply resample.

Before moving on, there is an issue in choosing $\alpha$. If it is set too low, certain hypotheses that are actually less consistent with the data will look more plausible than the true RM. We must set $\alpha$ sufficiently high to assign most of our probability mass on a good region of reward space. However, if $\alpha$ is set too high, this makes it difficult to move through reward space. Even rather similar hypotheses may have extremely different likelihoods, trapping us at local optima.

This is where using temperature helps. By raising the likelihood ratio to $1 / T_i$, we ``smooth out'' the differences between neighboring points. However, the likelihood ratio $P(\mathcal{D}\mid R_2) / P(\mathcal{D}\mid R_1)$ still takes on extreme values for large $\alpha$, making the prior ratio $P(R_2)/P(R_1)$ tiny by comparison, nullifying the utility of any prior information. We thus propose a modified transition rule, accepting $R_2$ over $R_1$ with probability: $\min \left(1, \left(\frac{P(\mathcal{D}\mid R_2)}{P(\mathcal{D}\mid R_1)}\right)^{1 / T_i} \frac{P(R_2)}{P(R_1)} \right)$. We can now set $\alpha$ high, use $T_i$ to get a more relaxed distribution, and maintain the importance of the prior information.


Repeat this process for $N$ iterations and output the best hypothesis seen. We choose the $R$ with the largest ``score'' $P(R\mid\mathcal{D})^{1 / T}P(R)$, smoothed by temperature $T$. We must use the same $T$ for all $R$ to make a consistent comparison, rather than using $T_i$ at the point where $R$ is sampled. We use the minimum value of the temperature schedule for $T$.

\subsection{Setting the Parameters and Schedules}
Setting $\alpha$ requires some tuning. If during trial runs the chosen RM $R$ does not fit the data very well, $\alpha$ should be raised. However, if $\alpha$ is set too high, there could be difficulty moving through reward space. The higher $\alpha$ is, and the more data we generate for the demonstration, the more extreme $P(\mathcal{D} \mid R_2) / P(\mathcal{D} \mid R_1)$ will tend to be. Accordingly, a higher initial temperature $T_0$ will need to be chosen, with a more gradual declining temperature schedule, to move smoothly through reward space.

We set our temperature schedules as follows. Set $T_0$ to some large initial value. Every $k$ steps, decay the temperature by constant factor $0 < \beta_T < 1$. Repeat until we hit minimum temperature $T$, at which point the temperature stays constant. $T$ should be chosen in conjunction with $\alpha$ such that we get a ``reasonable`` distribution over hypothesis space. Meaning, similar hypotheses usually do not have posteriors that differ by orders of magnitude, and prior information is still sometimes useful in deciding between hypotheses.
 
We follow a similar pattern for the perturbance probability schedule. Set $p_0$ to some middling initial value. Every $k$ steps, decay by constant factor $0 < \beta_p < 1$ to a minimum value of $p$, chosen so that some small number of transitions are still altered at each step (around $1$ or $2$ on average).

\subsection{Setting the Prior}
Finally, we must set the prior $P(R)$. In early experiments, we simply used a uniform prior. However, we found it useful to place a prior in favor of ``simple'' transitions. That is, a prior in favor of zero-reward, self-transitions.

For each transition, suppose it has prior probability $p_r$ of being zero-reward and $p_t$ of being a self-transition. The rest of the probability mass is split uniformly between the remaining possible rewards and states. Suppose $G = |\Gamma|$. If $R$ has $i$ zero-reward transitions and $j$ self-transitions, then its prior is $\tilde P(R) = p_r^i [(1-p_r)/(G-1)]^{nm-i} p_t^j [(1-p_t)/(n-1)]^{nm-j}$. Note that this is an \emph{unnormalized prior}; it does not take into account how many different RM structures there are. However, it suffices to calculate the prior ratio, which is all that is needed. Thus, $P(R_2)/P(R_1) = \tilde P(R_2)/\tilde P(R_1)$.

\section{Related Work}
\label{sec:relwork}
As mentioned in the introduction, approaches to IRL include margin optimization \cite{Ng00algorithmsfor} \cite{Abbeel04apprenticeshiplearning} \cite{Neu2007}, maximum entropy \cite{ziebart2008maximum} \cite{Wulf2015} \cite{Boularias2012}, and Bayesian \cite{Ramachandran2007} \cite{Lopes09activelearning} \cite{NIPS2011_c51ce410} methods. All of these infer a Markovian reward function. Recent work considered non-Markovian RL using RMs \cite{ToroIcarte2020}, including inferring such RMs \cite{Icarte2021} \cite{Xu2019} \cite{Xu2020}. However, these methods require access to the reward signal, which we do not assume.

More closely related are the works of \cite{Memarian_2020} and \cite{Vazquez2018}. In the former, the authors develop a deep maximum entropy IRL algorithm for ``temporally extended tasks", specified by a deterministic finite automaton (DFA). A series of observed ``subgoals" is either accepted as satisfying the overall goal, or otherwise rejected. They apply the $L^*$ algorithm \cite{angluin1987learning} to learn the DFA structure and then infer a reward function over the joint states of the product MDP/DFA. However, this work operates with a functionally binary reward ($1$ for achieving the goal, $0$ otherwise). Then, a ``teacher'' answers the learner's queries about which sequences satisfy the goal. This implicitly gives reward information to the agent. In contrast, our algorithm applies with \textit{no} reward information. Our algorithm can also learn non-binary rewards.

In the latter work, the authors also learn a task specification, which they interpret as a Boolean non-Markovian reward. In contrast to the former work, they manage to do this purely from expert behavior. Their method has a lot of complexity to it, and in any case still only learns binary rewards, in contrast to our method.

\section{Experiments}
\label{sec:exp}
We now test our algorithm on a set of gridworlds. \citet{Memarian_2020} showed that applying standard memoryless IRL when the true reward signal is non-Markovian gives poor results, often resulting in no reward at all. Therefore, a more sophisticated method is certainly necessary.

As mentioned before, the method in \cite{Memarian_2020} does not apply in our setting, since it requires more information than just expert behavior. \citet{Vazquez2018} is the only work we know of which solves roughly the same problem we do, to infer a non-Markovian reward directly from the expert demonstration. As such, it may seem natural to compare against their method. In fact, we use the running example from their paper as one of our gridworlds.

However, this latter method is prohibitively slow in our setting. It requires first defining a ``concept class'', i.e. the hypothesis space. Just to set up the problem, it is necessary to determine the subset relationships between \emph{all pairs of hypotheses}, in terms of the language accepted by each binary specification (of course, many pairs are incomparable). A lattice structure is then built from the hypothesis space and operated on to find the best specification. \citet{Vazquez2018} do this in a reasonable amount of time by considering a concept space of only 930 specifications. In our setting, we allow all possible RMs of a fixed size and end up with millions of specifications, even for the small problems we are considering. This results in trillions of pairs, which takes weeks or months to deal with. As such, this method is not suitable for our setting.

We consider three gridworlds. The first is from \cite{Vazquez2018}. There, the expert attempts to reach a recharging station while avoiding lava and water. If the expert does get wet, it can navigate to a towel to dry off before recharging. The second is from \cite{ToroIcarte2020}, in which the expert attempts to deliver a cup of coffee to the office, while avoiding any decorations. Finally, we create a modification of the coffee gridworld where one of the machines produces weak coffee, yielding a reward of $1$ when brought to the office, and the other produces strong coffee, yielding a reward of $2$.



In all cases, we set $\alpha = 50$. We also set the transition prior as $p_t = 3/5$ and the reward prior as $p_r = 3/4$. The values of the other parameters are shown in Table 1. In each example, we perform three independent runs of the algorithm and keep the highest scoring hypothesis. We then train the learning agent on this inferred reward function using policy iteration and finally run the learned policy on the given gridworld for $100$ episodes. In each episode, we sum the rewards achieved by the agent at each step, then compute an average over the $100$ episodes. We do the same for the expert and compare. The final results can be seen in Table 1.

For both clarity and reminder, in Table 1, $n$ denotes how many states are in the true reward machine (and is thereby the size of each hypothesis considered), ``runs'' and ``ep\_len'' are the number and length of episodes used to generate the demonstration, respectively, $N$ is the number of samples we generate in the simulated annealing process, and $r_e$ and $r_a$ are the average expert and agent reward, respectively.

The learner performs as well as or better than the expert on these gridworlds. The learner can do better because we make the expert noisy, while we allow the agent to fully optimize the inferred reward function. If the agent is able to infer an accurate reward function from the noisy expert, it can outperform the given demonstration.

\begin{table}[tb]
  \label{tab:exp}
  \begin{center}
  \begin{tabular}{|r|r|r|r|}
    \hline
    \textit{} & \textit{Recharge} & \textit{Coffee} & \textit{Multi Coffee} \\
    \hline
    $n$ & 3 & 3 & 4 \\
    \hline
    runs & 1,000 & 100 & 300 \\
    \hline
    ep\_len & 25 & 100 & 100 \\
    \hline
    $N$ & 2,000 & 1,000 & 10,000\\
    \hline
    $T_0$ & 500,000 & 100,000 & 1,000,000 \\
    \hline
    $T$ & 200 & 300 & 50 \\
    \hline
    $\beta_T$ & 0.98 & 0.96 & 0.99 \\
    \hline
    $p_0$ & 0.5 & 0.5 & 0.5 \\
    \hline
    $p$ & 1/16 & 1/12 & 1/16 \\
    \hline
    $\beta_p$ & 0.99 & 0.99 & 0.995 \\
    \hline
    $k$ & 5 & 5 & 10 \\
    \hline
    $r_e$ & 1.0 & 0.55 & 1.37 \\
    \hline
    $r_a$ & 1.0 & 0.68 & 2.0 \\
    \hline
  \end{tabular}
  \end{center}
  \caption{Experimental Results}
\end{table}

\section{Conclusion and Future Work}
\label{sec:conc}
We have devised an algorithm for extending existing BIRL techniques to the domain of non-Markovian reward functions. We define our hypothesis space to be the space of RMs, project our demonstration onto the state space of each hypothesis product MDP $M \times R$, and apply simulated annealing to find a MAP estimator for the posterior distribution. There are several avenues for future work.

First, concerning the algorithm, there could be a more efficient optimization method than simulated annealing. Other variants could make use of simulated tempering \cite{Marinari_1992}, parallel tempering \cite{Earl_2005}, beam search, or any other discrete optimization method.

Further, since we only need Q-values for the state-action pairs found in $\mathcal{D}$, it would be more efficient to obtain an estimate of the Q function only on $\mathcal{D}$. Potential algorithms include prioritized sweeping, trajectory sampling, and real-time dynamic programming. It is not immediately obvious, however, how they would map onto the problem at hand.

Finally, there are methods introduced in \cite{Michini_2012} for improving the efficiency of BIRL. A kernel function is introduced to measure similarity between states. When updating the guess for the reward vector, states are sampled according to their similarity to those in the demonstration. While these methods could improve efficiency, we need to understand how to apply such methods with the expansion of our state space.  

For conceptual advancements, our method is limited by assuming the number of states has a known fixed value of $n$. It would be preferable if $n$ was itself a learned parameter. However, this would make the sizes of our reward and transition matrices change as $n$ changes. Thus, the dimensionality of our hypothesis space would change as we move through it. The method of ``reversible jump'' MCMC \cite{Green_1995} allows both the number of parameters and their values to change, which could possibly be adapted to our current setting.

Finally, the entire method is a passive one. There has been work in using active methods, where the learner issues \textit{queries} to the expert, to enhance learning \cite{Lopes09activelearning}. The idea is to query states for which the learner has maximum entropy, and thus will learn the most by seeing the expert's action. An avenue of future investigation would be to extend these methods to the current setting.

\section*{Acknowledgments}
This work was supported in part by NSF Award CCF-2106339 and NSF CAREER Award CCF-1552497.

\bibliography{refs}

\title{Supplementary Material}
\author{}
\maketitle
\section{Experiments - Supplementary}
\label{sec:expsup}
\begin{figure}[tb]
  \centering
  \includegraphics[width=0.7\linewidth]{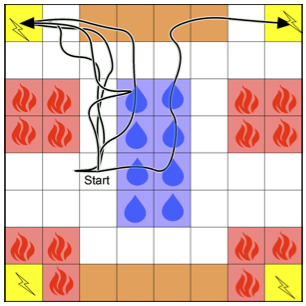}
  \caption{Recharge gridworld environment from \cite{Vazquez2018}}
  \label{fig:recharge}
\end{figure}

\begin{figure}[tb]
  \centering
  \includegraphics[width=0.8\linewidth]{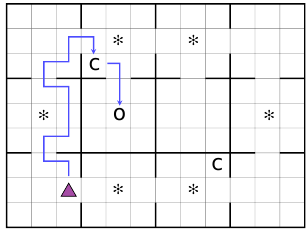}
  \caption{Office gridworld environment from \cite{ToroIcarte2020}}
  \label{fig:coffee}
\end{figure}

The gridworlds used in the main paper can be seen in Figures \ref{fig:recharge} and \ref{fig:coffee}. Recall that the third gridworld is a modification of the coffee gridworld where one of the machines produces weak coffee, yielding a reward of $1$ when brought to the office, and the other produces strong coffee, yielding a reward of $2$. The original coffee world is stochastic; on each move, the agent has a $10\%$ chance of slipping orthogonal to the intended direction ($5\%$ for each direction). The remaining two gridworlds are deterministic.

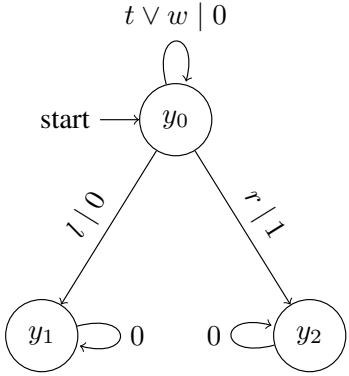
\begin{figure}[tb]
    \centering
    \resizebox{!}{0.6\linewidth}{
        \begin{tikzpicture}
        \node[state,initial] (0) {$y_0$};
        \node[state] (1) [below left=2cm and 1cm of 0] {$y_1$};
        \node[state] (2) [below right=2cm and 1cm of 0] {$y_2$};

        \path[->] (0) edge[loop above] node[align=center] {$t \vee w \mid 0$} (0);
        \path[->] (0) edge node[sloped,above] {$l \mid 0$} (1);
        \path[->] (0) edge node[sloped,above] {$r \mid 1$} (2);
        
        \path[->] (1) edge[loop right] node[align=center] {$0$} (1);
        
        \path[->] (2) edge[loop left] node[align=center] {$0$} (2);
        \end{tikzpicture}
    }
    \caption{Inferred reward machine for recharge gridworld.}
    \label{fig:rm1}
\end{figure}

\begin{figure}[tb]
    \centering
    \resizebox{!}{1.05\linewidth}{
        \begin{tikzpicture}
        \node[state,initial] (0) {$y_0$};
        \node[state] (1) [below left=2cm and 1cm of 0] {$y_1$};
        \node[state] (2) [below right=2cm and 1cm of 0] {$y_2$};
        \node[state] (3) [below=4cm of 0] {$y_3$};

        \path[->] (0) edge[loop above] node[align=center] {$o \mid 0$} (0);
        \path[->] (0) edge node[sloped,below] {$c \mid 0$} (1);
        \path[->] (0) edge node[sloped,below] {$k \mid 0$} (2);
        \path[->] (0) edge[bend right=0.5cm] node[sloped,below] {$* \mid 0$} (3);
        
        \path[->] (1) edge[loop above] node[align=center] {$c \vee k \mid 0$} (1);
        \path[->] (1) edge node[above,below] {$* \mid 0$} (2);
        \path[->] (1) edge node[below,sloped] {$o \mid 2$} (3);
        
        \path[->] (2) edge[loop above] node[align=center] {$c \vee k \mid 0$} (2);
        \path[->] (2) edge node[above,sloped] {$* \mid 0$} node[below,sloped] {$o \mid 1$} (3);
        
        \path[->] (3) edge[loop below] node[align=center] {$0$} (3);
        \end{tikzpicture}
    }
    \caption{Inferred reward machine for multi coffee gridworld.}
    \label{fig:rm3}
\end{figure}
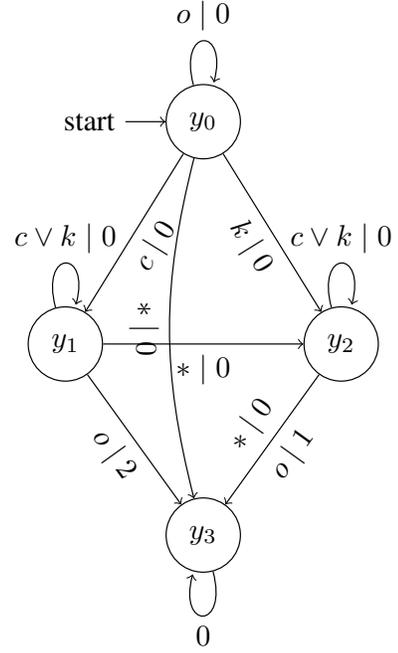

We can analyze the inferred reward functions qualitatively by looking at Figures \ref{fig:rm1}, \ref{fig:rm2}, and \ref{fig:rm3}. In each, an arrow between two states with label $\ell \mid r$ indicates that we transition from the first state to the second on input symbol $\ell$ and receive reward $r$. We use $\vee$ to indicate multiple symbols that cause the same transition. If we leave out $\ell$ entirely, this indicates that all input symbols yield the same transition.

Figure \ref{fig:rm1} is the inferred reward machine for the recharge gridworld; $t$ corresponds to the drying towel, $w$ to the water, $l$ to the lava, and $r$ to the recharging station. The agent has correctly inferred that reaching the recharge station is rewarding while touching lava is not. Unfortunately, it has failed to infer that recharging while wet is \emph{not} rewarding. The expert rarely exhibits this behavior so it does not have much weight in the demonstration. The learning agent is still able to perform well, however, since there is no water on the optimal path.

Figure \ref{fig:rm2} is the inferred reward machine for the basic coffee gridworld; $c$ corresponds to the coffee, $o$ to the office, and $*$ to the decorations. The learner has in fact inferred exactly the correct reward machine.

Finally, Figure \ref{fig:rm3} is the inferred reward machine for the multi coffee gridworld. Here $c$ corresponds to the strong coffee and $k$ to the weak. This machine is almost exactly correct. The only error is that the agent believes that if it observes $*$ from state $y_1$ (in which it is holding the strong coffee), it transitions to state $y_2$ (which should correspond to holding the weak coffee). Rather, it should transition to the terminal state $y_3$. This does not affect optimal behavior, since on this hypothesis, the agent should still pick up the strong coffee and avoid stepping on any decorations. Notably, the agent correctly infers the non-binary structure of the reward function, learning that the strong coffee yields a reward of $2$ but the weak coffee only $1$.

\begin{figure}[tb]
    \centering
    \resizebox{!}{0.65\linewidth}{
        \begin{tikzpicture}
        \node[state,initial] (0) {$y_0$};
        \node[state] (1) [below left=2cm and 1cm of 0] {$y_1$};
        \node[state] (2) [below right=2cm and 1cm of 0] {$y_2$};

        \path[->] (0) edge[loop above] node[align=center] {$o \mid 0$} (0);
        \path[->] (0) edge node[sloped,above] {$c \mid 0$} (1);
        \path[->] (0) edge node[sloped,above] {$* \mid 0$} (2);
        
        \path[->] (1) edge[loop above] node[align=center] {$c \mid 0$} (1);
        \path[->] (1) edge node[above,sloped] {$* \mid 0$} node[below,sloped] {$o \mid 1$} (2);
        
        \path[->] (2) edge[loop above] node[align=center] {$0$} (2);
        \end{tikzpicture}
    }
    \caption{Inferred reward machine for basic coffee gridworld.}
    \label{fig:rm2}
\end{figure}
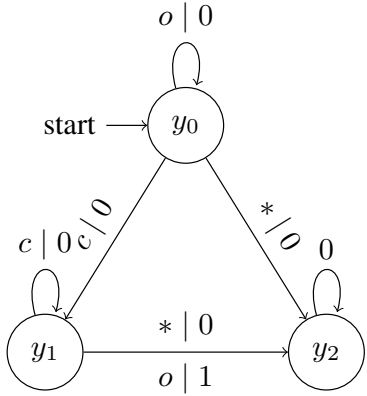

\section{Pseudocode}

\begin{algorithm}[tb]
\caption{Non-Markovian BIRL}
\label{alg:birl}

\textbf{Input}: expert demonstration $\mathcal{D}$\\
\textbf{Output}: estimate of maximum reward machine $R_{max}$

\begin{algorithmic}[1] 
\STATE $p := p_0$
\STATE $T := T_0$
\STATE Choose a valid RM $R_1$ uniformly at random
\STATE $R_{max} := R_1$
\STATE Run policy iteration on $M \times R_1$ to compute its optimal Q-values

\FOR{$1 \leq 1 \leq N$}
    \STATE Choose valid neighboring RM $R_2$ by independently updating the reward/target state of each transition of $R_1$ with probability $p$ to a new uniformly random value
            
    \STATE Run policy iteration on $M \times R_2$ to compute its optimal Q-values
    
    \STATE // Update highest scoring RM
    \IF{$\left(\frac{P(\mathcal{D}\mid R_2)}{P(\mathcal{D}\mid R_{max})}\right)^{1 / T_{min}} \frac{P(R_2)}{P(R_{max})} > 1$}
        \STATE $R_{max} := R_2$
    \ENDIF
            
    \STATE With probability $\min \left(1, \left(\frac{P(\mathcal{D}\mid R_2)}{P(\mathcal{D}\mid R_1)}\right)^{1 / T} \frac{P(R_2)}{P(R_1)} \right)$, set $R_1 := R_2$
    
    \STATE // Update parameters every $k$ steps
    \IF{$i \mod k = 0$}
        \STATE $T := \max(T \cdot \beta_T, T_{min})$
        \STATE $p := \max(p \cdot \beta_p, p_{min})$
    \ENDIF
\ENDFOR

\STATE \textbf{return} $R_{max}$
\end{algorithmic}
\end{algorithm}

For the sake of clarity, we have included pseudocode for our algorithm in Algorithm \ref{alg:birl}.

\end{document}